\documentclass[letterpaper, 10 pt, journal, twoside]{IEEEtran}

\usepackage{decar-common}
\usepackage{decar-dynamics}
\usepackage{decar-lie}
\usepackage{decar-post}
\usepackage{tabularx}
\usepackage{booktabs}

\DeclareMathOperator{\blkdiag}{blkdiag}

\newlength{\subfigheight}
\newsavebox{\subfigbox}

\Crefformat{figure}{#2Fig.~#1#3}

\makeatletter
\AtBeginDocument{
  \check@mathfonts
}
\makeatother

\newcolumntype{C}[1]{>{\centering\let\newline\\\arraybackslash\hspace{0pt}}m{#1}}
\newcolumntype{Y}{>{\centering\arraybackslash}X}

\newcommand{\update}[1]{\colour{black}{#1}}

\begin{document}

%
%
%
%
%
%
%
\def \myJournal {IEEE Robotics and Automation Letters}
\def \myDoi {10.1109/LRA.2024.3367270}
\def \myPaperSiteName {IEEE Xplore}
\def \myPaperSiteLink {https://ieeexplore.ieee.org/document/10439640}
\def \myYear {2024}
\def \myPaperCitation{T. Hitchcox and J. R. Forbes, ``Laser-to-Vehicle Extrinsic
Calibration in Low-Observability Scenarios for Subsea Mapping,'' in \textit{IEEE
Robotics and Automation Letters}, vol. 9, no. 4, pp. 3522-3529, 2024.}


\begin{figure*}[t]

\thispagestyle{empty}
\begin{center}
\begin{minipage}{6in}
\centering
This paper has been accepted for publication in \emph{\myJournal}. 
\vspace{1em}

This is the author's version of an article that has, or will be, published in this journal or conference. Changes were, or will be, made to this version by the publisher prior to publication.
\vspace{2em}

\begin{tabular}{rl}
DOI: & \myDoi\\
\myPaperSiteName: & \texttt{\myPaperSiteLink}
\end{tabular}

\vspace{2em}
Please cite this paper as:

\myPaperCitation

\vspace{15cm}
\copyright \myYear \hspace{4pt}IEEE. Personal use of this material is permitted. Permission from IEEE must be obtained for all other uses, in any current or future media, including reprinting/republishing this material for advertising or promotional purposes, creating new collective works, for resale or redistribution to servers or lists, or reuse of any copyrighted component of this work in other works.

\end{minipage}
\end{center}
\end{figure*}
\newpage
\clearpage
\pagenumbering{arabic} 

\fontdimen16\textfont2=\fontdimen17\textfont2
\fontdimen13\textfont2=5pt

\title{Laser-to-Vehicle Extrinsic Calibration in Low-Observability Scenarios for Subsea Mapping}

\author{Thomas~Hitchcox,~\IEEEmembership{Member,~IEEE,}
        and~James~Richard~Forbes,~\IEEEmembership{Member,~IEEE}%
        
\thanks{Manuscript received: September 28, 2023; Revised: December 17, 2023;
Accepted: February 3, 2024. This paper was recommended for publication by Editor
Giuseppe Loianno upon evaluation of the Associate Editor and Reviewers’
comments.  This work was supported by Voyis Imaging Inc. through the Natural
Sciences and Engineering Research Council of Canada (NSERC) Collaborative
Research and Development (CRD) program, and the McGill Engineering Doctoral
Award (MEDA) program.}%

\thanks{Thomas~Hitchcox and James~Richard~Forbes are with the Department of Mechanical
Engineering, McGill University, Montreal, Quebec H3A~0C3, Canada.
\texttt{thomas.hitchcox@mail.mcgill.ca, james.richard.forbes@mcgill.ca}.}%

\thanks{Digital Object Identifier (DOI): see top of this page.}}%

\markboth{IEEE Robotics and Automation Letters. Preprint version. Accepted
February, 2024}{Hitchcox and Forbes: Laser-to-Vehicle Extrinsic Calibration in Low-Observability Scenarios for Subsea Mapping}%

\maketitle

\begin{abstract}
    Laser line scanners are increasingly being used in the subsea industry for
    high-resolution mapping and infrastructure inspection.  However, calibrating
    the 3D pose of the scanner relative to the vehicle is a perennial source of
    confusion and frustration for industrial surveyors.  This work describes
    three novel algorithms for laser-to-vehicle extrinsic calibration using
    naturally occurring features.  Each algorithm makes a different assumption
    on the quality of the vehicle trajectory estimate, enabling good calibration
    results in a wide range of situations.  A regularization technique is used
    to address low-observability scenarios frequently encountered in practice
    with large, rotationally stable subsea vehicles.  Experimental results are
    provided for two field datasets, including the recently discovered wreck of
    the \textit{Endurance}.  
\end{abstract}


\section{Introduction}
\label{sec:intro}

\IEEEPARstart{S}{ensor-to-vehicle} extrinsics define the 3D pose of a sensor
relative to the vehicle.  An accurate extrinsic estimate is critical for mapping
applications, as data collected from the sensor must be resolved in a common
reference frame using the vehicle's estimated trajectory.  Errors in the
extrinsic estimate will therefore have a direct impact on map quality.  

The subsea industry is increasingly using laser line scanners to produce
millimeter-resolution reconstructions of underwater assets.  To accurately
assess potential damage to these assets, the pose of the laser relative to the
vehicle must be known to within tenths of a degree and fractions of a
centimeter.  However, calibrating the laser-to-vehicle extrinsics is currently a
challenge in the subsea industry.  Computer-aided design (CAD) models of an
underwater vehicle and sensor payload should provide a good initial pose
estimate, however as-designed and as-deployed configurations can differ due to
in-field modifications.  ``Patch test'' operations are performed to refine an
extrinsic estimate, where a vehicle makes multiple passes over the same patch of
seabed, typically following prescribed maneuvers to isolate the effects of
individual degrees of freedom.  Patch tests are long and tedious, with extrinsic
refinement often performed manually to achieved a visually pleasing map.  This
paper delivers a straightforward, optimization-based approach to address this
problem.

\begin{figure}[t]
	\sbox\subfigbox{%
	  \resizebox{\dimexpr0.98\columnwidth-1em}{!}{%
		\includegraphics[height=4cm]{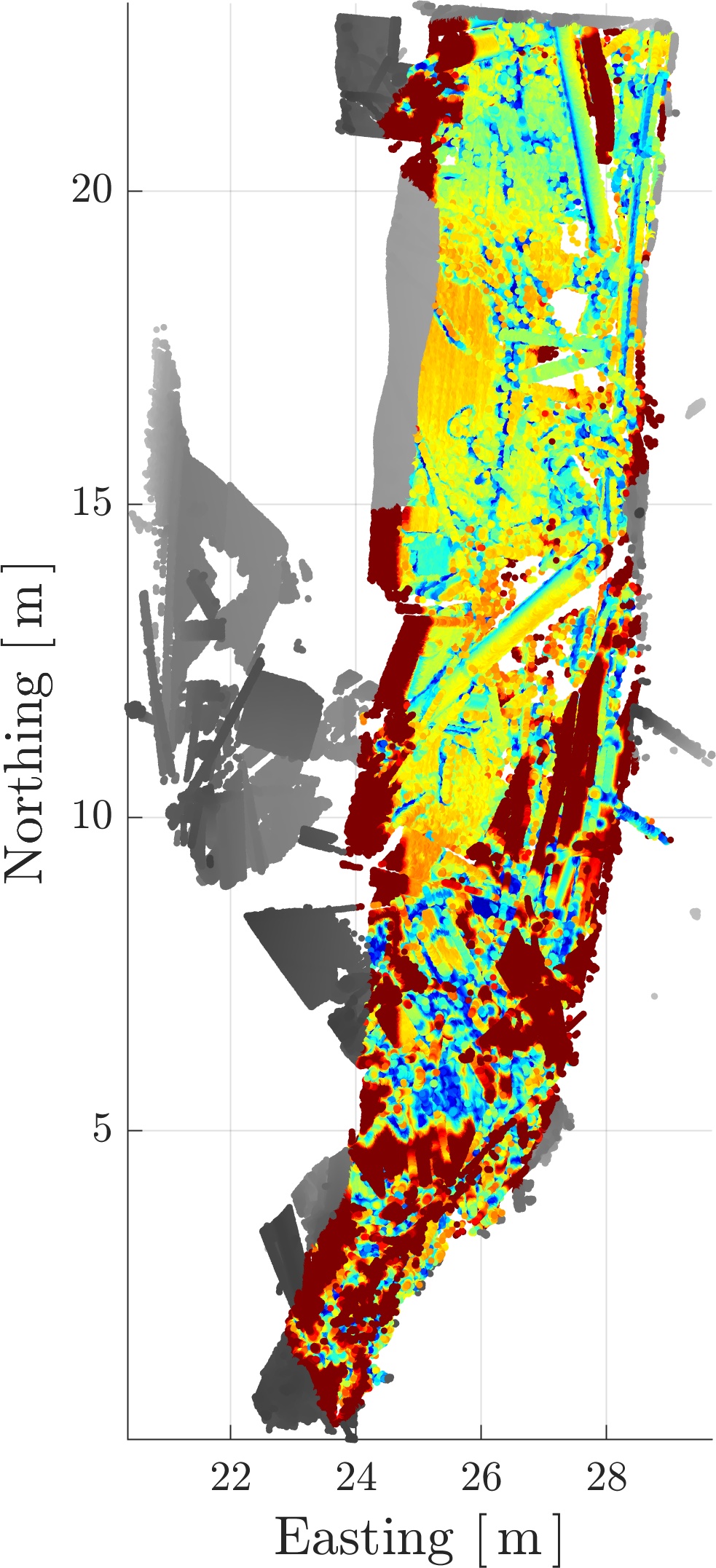}%
		\includegraphics[height=4cm]{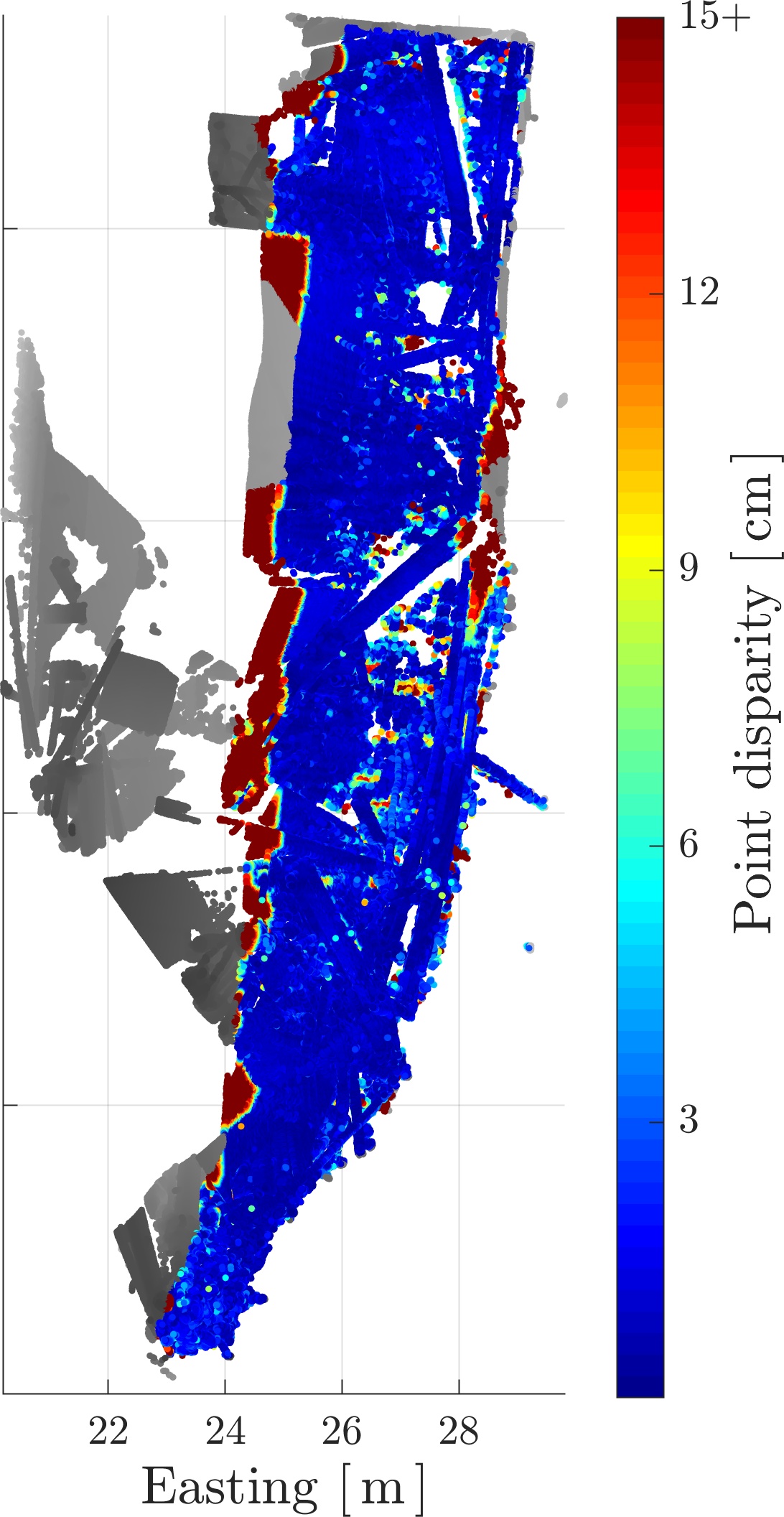}%
	  }%
	}
	\setlength{\subfigheight}{\ht\subfigbox}
	\centering
	\subcaptionbox{Prior \label{fig:extrinsics_endurance_prior}}{%
	  \includegraphics[height=\subfigheight]{figs/endurance_prior.jpg}
	}%
    \hspace{3pt}%
	\subcaptionbox{Posterior, Algorithm 2 \label{fig:extrinsics_endurance_posterior}}{%
	  \includegraphics[height=\subfigheight]{figs/endurance_posterior.jpg}
	}
    \caption{Patch test scans of the \textit{Endurance} shipwreck, with point
    disparity errors \cite{Roman2006} shown before and after joint extrinsic and
    trajectory optimization.  Included with permission from the Falklands
    Maritime Heritage Trust.}
    \label{fig:extrinsics_endurance}
\end{figure}

Sensor-to-vehicle extrinsic calibration has been extensively studied in the
literature. A general theory for spatial and temporal calibration is given in
\cite{Furgale2013}, which uses a continuous-time trajectory representation based
on B-splines.  A second general calibration framework presented in
\cite{Taylor2016} uses a maximum-likelihood approach, with refinement through
the minimization of various appearance-based metrics.  

Few studies consider laser profile scanners, however \cite{Napier2013} provides
a solution for laser-to-camera calibration. Studies investigating lidar-to-IMU
calibration are much more common. For example, \cite{Maddern2012} provides a
method for calibrating lidar-to-INS extrinsics by maximizing an entropy metric,
which represents the compactness of the resulting point cloud.  Online
lidar-to-IMU calibration is performed in \cite{Ye2019a} as part of a
lidar-inertial sliding window odometry framework. A continuous-time B-spline
trajectory representation is used for targetless lidar-to-IMU calibration in
\cite{Lv2020}.

During sensor extrinsic calibration, certain degenerate motion patterns result
in poor observability of calibration parameters. These patterns are identified
in \cite{Mirzaei2012,Taylor2016,Zuo2020}, with the need for roll and pitch
excitation for extrinsic calibration on ground vehicles identified in
\cite{Maddern2012}.  An observability-aware approach for extrinsic calibration
is developed in \cite{Lv2022a}, which uses singular value decomposition to
identify good sections of calibration data and to modify the state update during
optimization.  The same observability-aware approach to data selection is used
in \cite{Das2023}, which bounds the magnitude of the state update assuming that
a good initial estimate is available.

This work develops three novel algorithms for calibrating the pose of a laser
line scanner to a vehicle's inertial navigation system (INS).  The algorithms
calibrate all six degrees of freedom simultaneously by optimizing over the
matrix Lie group $SE(3)$, eliminating the need for complicated patch test
maneuvers and manual tuning.  Calibration is performed using naturally-occurring
3D features without the need for a reference target, and has been adapted for
use with rotationally stable underwater vehicles.  The algorithms are validated
on two challenging underwater field datasets, including a 3D laser
reconstruction of the historic \textit{Endurance} shipwreck
(\Cref{fig:extrinsics_endurance}).

\section{Preliminaries}
\label{sec:preliminaries}

\subsection{Reference Frames and Navigation Conventions}
\label{sec:frames}

This section discusses the reference frames and navigation conventions used in
this paper.  A three-dimensional dextral reference frame $\rframe{a}$ is
composed of three orthonormal basis vectors.  The position of point $z$ relative
to point $w$ is resolved in $\rframe{a}$ as ${\mbf{r}^{zw}_a \in \rnums^3}$ and
in frame $\rframe{b}$ as $\mbf{r}^{zw}_b$.  These quantities are related via
${\mbf{r}^{zw}_a = \mbf{C}_{ab} \mbf{r}^{zw}_b}$, with $\mbf{C}_{ab}$ a
direction cosine matrix, ${\mbf{C} \in SO(3) = \{ \mbf{C} \in \rnums^{3\times3}
\, | \, \mbf{C}\mbf{C}^\trans = \eye, \det \mbf{C} = +1 \}}$
\cite[\textsection7.1.1]{Barfoot2017}.  Time-varying quantities are indicated by
the subscript $(\cdot)_k$, for example $\mbf{r}^{z_kw}_a$ describes the position
of moving point $z$ at time $t_k$.  In this work $\rframe{a}$ is the local
geodetic navigation frame, $\rframe{b}$ is the vehicle frame, and
$\rframe{\ell}$ is the reference frame of the laser.  Following maritime
convention, $\rframe{a}$ and $\rframe{b}$ are north-east-down (NED) reference
frames, while the east-north-up (ENU) reference frame of the laser is shown in
\Cref{fig:insightpro}.  Finally, point $w$ is the world datum, point $z$ is the
vehicle datum, and point $s$ is the laser datum.

\subsection{Matrix Lie Groups}
\label{sec:liegroups}

\update{The 3D pose of a vehicle at time $t_k$ is represented as an element of
the matrix Lie group $SE(3)$,}
\begin{equation}
    \mbf{T}^{z_kw}_{ab_k} = \begin{bmatrix}
        \mbf{C}_{ab_k} & \mbf{r}^{z_kw}_a \\ \mbf{0} & 1 
    \end{bmatrix}
    \in SE(3),
    \label{eqn:pose}
\end{equation}
with ${SE(3) = \{ \mbf{T} \in \rnums^{4\times 4} \, | \, \mbf{C} \in SO(3),
\mbf{r} \in \rnums^3 \}}$ \cite[\textsection7.1.1]{Barfoot2017}.  In estimation
problems involving matrix Lie groups, perturbations and uncertainty are modeled
in the matrix Lie algebra ${\mathfrak{se}(3) \triangleq T_\eye SE(3)}$
\cite{Sola2018}, with ${\mbs{\xi}^\wedge \in \mathfrak{se}(3)}$.  The ``wedge''
operator is ${(\cdot)^\wedge : \rnums^6 \to \mathfrak{se}(3)}$, while the
``vee'' operator is ${(\cdot)^\vee : \mathfrak{se}(3) \to \rnums^6}$, such that
${(\mbs{\xi}^\wedge)^\vee = \mbs{\xi}}$.  A Lie group and Lie algebra are
related by the exponential map, which for matrix Lie groups is the matrix
exponential,
\vspace{-3pt}
\begin{equation}
    \mbf{T} = \exp(\mbs{\xi}^\wedge).
    \vspace{-3pt}
\end{equation}
The matrix logarithm is used to return to the Lie algebra via 
\vspace{-3pt}
\begin{equation}
    \mbs{\xi}^\wedge = \log(\mbf{T}).
    \vspace{-3pt}
\end{equation}
\indent
Errors on matrix Lie groups are defined multiplicatively.  This work uses a
left-invariant error definition \cite[\textsection2.3]{Arsenault2019},
\vspace{-3pt}
\begin{equation}
    \delta \mbf{T} = \mbf{T}\inv \mbftilde{T},
    \label{eqn:lefterror}
    \vspace{-3pt}
\end{equation}
where $\mbf{T}$ is the current state estimate and $\mbftilde{T}$ is a state
estimate generated from a predictive model or prior information.  The
corresponding perturbation scheme is
\begin{equation}
    \mbf{T} = \mbfbar{T} \exp(-\delta \mbs{\xi}^\wedge),
    \label{eqn:perturbation}
\end{equation}
with perturbation ${\delta \mbs{\xi} \sim \mathcal{N}(\mbf{0}, \mbs{\Sigma})}$,
${\mbs{\Sigma} = \expect{ \, \delta \mbs{\xi} \, \delta \mbs{\xi}^\trans} \in
\rnums^{6 \times 6}}$.  The state estimate is therefore defined by mean estimate
$\mbfbar{T}$ and covariance $\mbs{\Sigma}$.  Finally, this work makes use of the
$(\cdot)^\odot$ operator, ${(\cdot)^\odot : \rnums^4 \to \rnums^{4\times 6}}$
\cite[\textsection7.1.8]{Barfoot2017}, defined here for homogeneous point
${\mbf{u} = [ \, \mbf{r}^\trans \ 1 ]^\trans}$, ${\mbf{r} \in \rnums^3}$ as
\begin{equation}
    \mbf{u}^\odot = \begin{bmatrix}
        -\mbf{r}^\times & \eye \\ \mbf{0} & \mbf{0}
    \end{bmatrix},
    \label{eqn:odot}
\end{equation}
with $(\cdot)^\times$ the skew-symmetric operator
\cite[\textsection7.1.2]{Barfoot2017}.  

\subsection{Batch State Estimation}
\label{sec:batchestimation}

Given a set of states ${\mbc{X} = \{ \mbf{x}_i \}^N_{i=1}}$ to be estimated, a
set of measurements ${\mbc{Y} = \{ \mbf{y}_j \}^M_{j=1}}$ relating to the
states, and prior estimates ${\mbc{Z} = \{ \mbf{z}_i \}^N_{i=1}}$ of the states,
batch state estimation seeks to produce a maximum a posteriori (MAP) solution, 
\vspace{-3pt}
\begin{equation}
    \mbc{X}_\star = \argmax_{\mbc{X}} \, p \left( \mbc{X} \, | \, \mbc{Y}, \mbc{Z} \right).
    \label{eqn:map}
    \vspace{-3pt}
\end{equation}
\update{Invoking Bayes' rule, assuming Gaussian measurement noise densities, and
taking the negative log likelihood transforms \eqref{eqn:map} into a new
optimization problem \cite[\textsection3.1.2]{Barfoot2017},}
\vspace{-3pt}
\begin{equation}
    \mbc{X}_\star = \argmin_{\mbc{X}} \, J (\mbc{X}),
    \label{eqn:optprob}
    \vspace{-3pt}
\end{equation}
in which the least-squares objective function $J (\mbc{X})$ is
\vspace{-3pt}
\begin{equation}
    J (\mbc{X}) = \frac{1}{2} \sum^N_{i=1} \| \mbf{e}_i \! \left( \mbf{z}_i, \mbf{x}_i \right) \! \|^2_{\mbf{P}_i\inv} + \frac{1}{2} \sum^M_{j=1} \| \mbf{e}_j \! \left( \mbf{y}_j, \mbc{X}_j \right) \! \|^2_{\mbf{M}_j\inv}.
    \label{eqn:objfun}
    \vspace{-3pt}
\end{equation}
Here, $\mbf{e}_i$ denotes the prior errors, $\mbf{e}_j$ denotes the measurement
errors, $\mbc{X}_j$ is the set of states involved in defining error $\mbf{e}_j$,
and $\| \mbf{e} \|^2_{\mbs{\Sigma}\inv}$ is the squared Mahalanobis distance,
\begin{equation}
    \| \mbf{e} \|^2_{\mbs{\Sigma}\inv} \triangleq \mbf{e}^\trans \mbs{\Sigma}\inv \mbf{e} \in \rnums_{\ge 0},
\end{equation}
with $\mbf{P}_i$, $\mbf{M}_j$ denoting the covariance on the prior and
measurement errors, respectively.  \Cref{eqn:optprob} is solved by repeatedly
linearizing objective function \eqref{eqn:objfun} about the current operating
point $\bar{\mbc{X}}$ and obtaining the local minimizing increment $\delta
\mbc{X}_\star$ using, for example, Gauss-Newton or Levenberg-Marquardt
\cite[\textsection4.3.1]{Barfoot2017}.  For a problem on matrix Lie groups, the
Gauss-Newton update is 
\begin{equation}
    \delta \mbs{\xi}_\star = -\big( \mbf{F}^\trans \mbf{W} \, \mbf{F} \big)\inv \mbf{F}^\trans \mbf{W} \, \mbf{e}, 
    \label{eqn:gauss_newton}    
\end{equation}
in which ${\delta \mbs{\xi}_\star = [ \, \delta \mbs{\xi}_1^\trans \ \ldots \
\delta \mbs{\xi}_N^\trans \, ]^\trans }$, $\mbf{e}$ contains the stacked
$\mbf{e}_i$ and $\mbf{e}_j$, $\mbf{W}$ is a sparse matrix with the
$\mbf{P}_i\inv$ and $\mbf{M}_j\inv$ on the main block diagonal, and Jacobian
$\mbf{F}$ contains the individual ${\mbf{F}_i = \pd{\mbf{e}_i}{\mbs{\xi}_i}
\big|_{\mbfbar{T}_i} \!}$ and ${\mbf{F}^i_j =
\pd{\mbf{e}_j}{\mbs{\xi}_i}\big|_{\mbfbar{T}_i} \!}$ arranged accordingly.  With
a left-invariant error definition \eqref{eqn:lefterror}, the $SE(3)$ states are
updated as
\vspace{-3pt}
\begin{equation}
    \mbf{T}_i \leftarrow \mbf{T}_i \exp \left( -\delta \mbs{\xi}_i^\wedge \right).
\end{equation}

\begin{table*}[t]
    \centering
    \caption{Three algorithms for laser-to-INS extrinsic calibration}
    \renewcommand{\arraystretch}{1.5}
    \begin{tabularx}{\textwidth}{C{0.75cm}|m{6.25cm}|X|C{1.25cm}|C{1.25cm}}
    \toprule
    Alg. & Assumptions & Design variables & Figure & Section \\
    \hline
        1 & Perfect DVL-INS navigation estimate & Laser-to-INS extrinsics &
        \Cref{fig:reprojection_errors_1} & \ref{sec:alg1} \\
        2 & Good local navigation, global submap drift & Laser-to-INS
        extrinsics, global submap poses & \Cref{fig:reprojection_errors_2} &
        \ref{sec:alg2} \\
        3 & Poor DVL-INS navigation estimate & Laser-to-INS extrinsics, submap
        shape & \Cref{fig:reprojection_errors_3} & \ref{sec:alg3} \\
	\bottomrule
    \end{tabularx}
	\label{tab:algs_overview}
\end{table*}

\subsection{Subsea Laser Scanning}
\label{sec:laserscanning}

\begin{figure}[b]
	\centering
	\includegraphics[width=0.99\columnwidth]{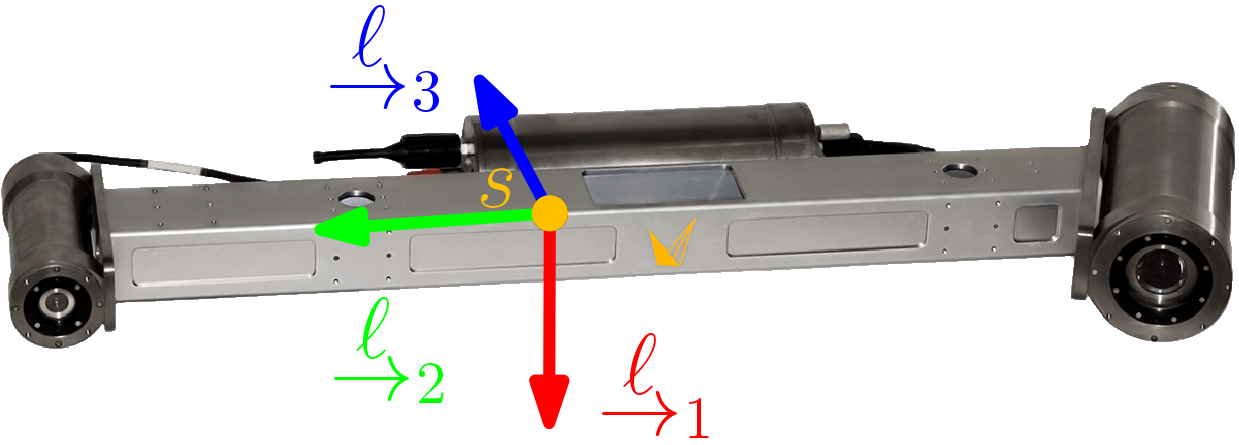}
	\caption{The Insight Pro underwater line scanner by Voyis Imaging Inc.,
	showing the approximate location of datum point $s$ and sensor reference
	frame $\rframe{\ell}$.  The baseline between the line projector (left) and
	the camera (right) is approximately \SI{1}{\meter}.}
	\label{fig:insightpro}
\end{figure}

Subsea laser line scanners, such as those developed by the University of Girona
\cite{Palomer2019a} and Voyis Imaging Inc. (\Cref{fig:insightpro}), use optical
triangulation to measure high-resolution 2D profiles of underwater environments.
A review of the state-of-the-art in underwater optical scanning may be found in
\cite{Castillon2019}.  This work considers the problem of laser-to-INS extrinsic
calibration using the Voyis Insight Pro laser scanner (\Cref{fig:insightpro}).
This sensor emits a pulsed laser swath with a beam angle of \SI{50}{\deg}, and
uses an optical camera to record 2D laser profiles at a frequency of up to
\SI{80}{\Hz}.  To generate 3D point-cloud ``submaps'' in the world frame, laser
measurements ${\mbf{r}^{p_is}_\ell \in \rnums^3}$ are registered to sections of
the vehicle trajectory via
\begin{equation}
    \begin{bmatrix}
        \mbf{r}^{p_iw}_a \\ 1
    \end{bmatrix} = \mbf{T}^{zw}_{ab} \, \mbf{T}^{sz}_{b\ell} \begin{bmatrix}
        \mbf{r}^{p_is}_{\ell} \\ 1
    \end{bmatrix},
    \label{eqn:registerprofiles}
\end{equation}
where extrinsic matrix ${\mbf{T}^{sz}_{b\ell} \in SE(3)}$ captures the static
pose of the laser relative to the INS.  Interpolating the vehicle trajectory at
measurement time $t_i$ is done via \cite[\textsection2.4]{Eade2014}
\begin{equation}
    \mbf{T}_i = \mbf{T}_k \exp \left( \left( \tfrac{t_i - t_k}{t_{k+1} - t_k} \right) \log \left( \mbf{T}\inv_k \mbf{T}_{k+1} \right) \right),
\end{equation}
with ${t_k < t_i < t_{k+1}}$ and, for example, ${\mbf{T}_i \leftarrow
\mbf{T}^{z_iw}_{ab_i}}$.  The resulting point-cloud submap is ${\mathcal{P} = \{
\mbf{r}^{p_iw}_a \}^P_{i=1} }$.

\section{Methodology}
\label{sec:methodology}

This section developes three novel algorithms for laser-to-vehicle extrinsic
calibration with a Doppler-velocity log (DVL)-aided INS (DVL-INS).  Reprojection
errors are first defined between pairs of 3D keypoints detected throughout $N$
laser scans.  Tikhonov regularization is then used to address low-observability
calibration scenarios often encountered with large, rotationally stable
underwater vehicles.  Three algorithms are derived for laser-to-INS extrinsic
calibration subject to different assumptions on the quality of the vehicle
trajectory estimate.  These assumptions are summarized in
\Cref{tab:algs_overview}.

Note that reliable keypoint extraction (\Cref{sec:keypoints}) depends heavily on
access to a reasonable prior laser-to-INS extrinsic estimate
$\mbfhat{T}^{sz}_{b\ell}$.  In practice this is a fair assumption, as a prior
estimate can almost always be obtained from a CAD model or even from on-site
photographs and hand measurements.  The certainty of these measurements can then
be quantified in the weighting placed on the Tikhonov regularization term.  This
is discussed in \Cref{sec:observability}.

\subsection{Extracting 3D Keypoints from Laser Submaps}
\label{sec:keypoints}

The calibration algorithms developed in \Cref{sec:threealgorithms} minimize a
sum of squared reprojection errors defined between pairs of 3D keypoints
detected in the $N$ laser scans.  To obtain a set of inlier matches
between the different scans, 3D SIFT keypoints \cite{Lowe2004}
\update{${\mbf{r}^{qw}_a \in \rnums^3}$} are detected and matched using the
TEASER++ coarse alignment algorithm \cite{Yang2020} on the basis of FPFH feature
correspondences \cite{Rusu2009a}.  \update{Failed alignments may be detected
though comparison to the INS attitude estimate.}  \Cref{tab:frontend_params}
provides a summary of the relevant parameters, \update{which were selected after
a small amount of tuning to produce good results on the structured datasets
studied in this work.} The result of this operation is a set of $M$ inlier
correspondences between pairs of detected keypoints, with each correspondence
allowing for the formulation of one reprojection error $\mbf{e}_j$.  The
formulation of this error is discussed in the next section.

\setlength{\tabcolsep}{4pt}
\begin{table}[b]
    \centering
    \caption{Keypoint detection and matching parameters}
    \label{tab:pcparams}
    \renewcommand{\arraystretch}{1.2}
    \begin{tabularx}{\columnwidth}{lX}
    \toprule
    Operation & Description \\ \hline
    Downsampling & Subsample on \SI{5}{\centi\meter} grid \\
    Normal vectors & 40 nearest neighbors \\ 
    Keypoints & 3D SIFT \cite{Lowe2004}, PCL v1.9 implementation \\
    Features & FPFH \cite{Rusu2009a}, PCL v1.9 implementation \\ 
    Feature matching & TEASER++ \cite{Yang2020}, 10 putative matches \\
    \bottomrule
    \end{tabularx}
    \label{tab:frontend_params}
\end{table}

\subsection{Defining and Minimizing Reprojection Errors}
\label{sec:reprojectionerrors}

\begin{figure}[t]
	\centering
	\includegraphics[width=\columnwidth]{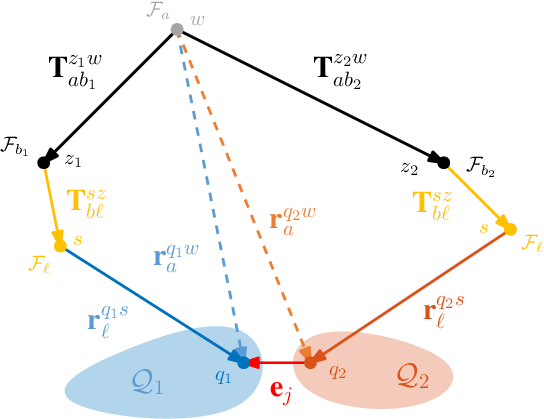}
	\caption{Defining a \colour{red}{reprojection error} between matched
	keypoints \colour{matlab_blue}{$\mbf{r}^{q_1w}_{a}$} and
	\colour{matlab_orange}{$\mbf{r}^{q_2w}_{a}$}.  Vehicle poses are shown in
	black.  The design variable is the \colour{voyis_yellow}{laser-to-INS
	extrinsics}.}
	\label{fig:reprojection_errors_1}
\end{figure}

Consider \Cref{fig:reprojection_errors_1}, in which keypoint submaps
$\mathcal{Q}_1$ and $\mathcal{Q}_2$ are generated according to
\eqref{eqn:registerprofiles} for two overlapping sections of the vehicle
trajectory.  For the $j^\textrm{th}$ inlier correspondence between keypoints
${\mbf{r}^{q_1w}_{a} \! \in \mathcal{Q}_1}$ and ${\mbf{r}^{q_2w}_{a} \! \in
\mathcal{Q}_2}$, the reprojection error is 
\begin{subequations}
    \begin{align*}
        \mbf{e}_j(\mbf{T}^{sz}_{b\ell})        =& \ \mbf{r}^{q_1w}_{a} - \mbf{r}^{q_2w}_{a} \\
                                               =& \ \mbf{H} \! \left( \mbf{T}^{z_1w}_{ab_1} \, \mbf{T}^{sz}_{b\ell} \, \mbf{u}^{q_1s}_\ell - \mbf{T}^{z_2w}_{ab_2} \, \mbf{T}^{sz}_{b\ell} \, \mbf{u}^{q_2s}_\ell \right),
        \label{eqn:rep_error_1} \numberthis \\
        \mbf{e}_j(\mbf{T}) =& \ \mbf{H} \big( \mbf{T}_1 \mbf{T} \, \mbf{u}_1 - \mbf{T}_2 \mbf{T} \, \mbf{u}_2 \big),
        \label{eqn:rep_error_2} \numberthis
    \end{align*}
\end{subequations}
in which ${\mbf{u} = [ \, \mbf{r}^\trans \ 1 ]^\trans}$ is the homogenous form
of ${\mbf{r} \in \rnums^3}$, ${ \mbf{H} = [ \, \eye \ \ \mbf{0} \,] \in
\rnums^{3 \times 4}}$ removes the homogenous component, and where the notation
is simplified from \eqref{eqn:rep_error_1} to \eqref{eqn:rep_error_2} such that,
for example, ${\mbf{T}_1 \leftarrow \mbf{T}^{z_1w}_{ab_1}}$, ${\mbf{T}
\leftarrow \mbf{T}^{sz}_{b\ell}}$, and ${\mbf{u}_1 \leftarrow
\mbf{u}^{q_1s}_\ell}$.

Next, the Jacobians ${\mbf{F}_j = \pd{\mbf{e}_j}{\mbs{\xi}}\big|_{\mbfbar{T}} }$ are obtained by
linearizing the error model.  Perturbing design variable $\mbf{T}$ in a
left-invariant sense, with ${\exp(-\delta \mbs{\xi}^\wedge) \approx (\eye -
\delta \mbs{\xi}^\wedge)}$, \eqref{eqn:rep_error_2} is linearized as 
\begin{equation}
	\mbf{e}_j = \mbfbar{e}_j + \underbrace{ \mbf{H} \! \left( \mbf{T}_2 \mbfbar{T} \, \mbfbar{u}_2^\odot - \mbf{T}_1 \mbfbar{T} \, \mbfbar{u}_1^\odot \right)}_{\mbf{F}_j} \! \delta \mbs{\xi} + \underbrace{\mbf{C}_1 \mbfbar{C}}_{\mbf{G}_j^1} \delta \mbf{r}_1 \underbrace{-\mbf{C}_2 \mbfbar{C}}_{\mbf{G}_j^2} \delta \mbf{r}_2,
	\label{eqn:extrinsics_perturb_epsilon}
\end{equation}
in which second-order terms have been ignored and where, for example,
${\mbf{r}_1 \leftarrow \mbf{r}^{q_1w}_{a}}$, ${\mbf{r}_1 = \mbfbar{r}_1 + \delta
\mbf{r}_1}$.  Letting, for example, ${ \mbf{R}_1 = \expect{ \, \delta \mbf{r}_1
\, \delta \mbf{r}_1^\trans \, } }$ represent the covariance on the first point
measurement, and assuming the point measurements are uncorrelated, the
covariance on the reprojection error is
\begin{equation}
    \mbf{M}_j = \mbf{G}^1_j \, \mbf{R}_1 (\mbf{G}^1_j)^\trans + \mbf{G}^2_j \, \mbf{R}_2 (\mbf{G}^2_j)^\trans.
    \label{eqn:extrinsics_sigma}
\end{equation}
\update{A sensitivity study on simulated data
\cite[\textsection5.3]{Hitchcox2023thesis} suggests that, for practical
applications, between six to ten point-cloud submaps are needed, with at least
twenty common keypoints identified in each submap.  This is easily achieved for
the structured shipwreck datasets studied in \Cref{sec:results}, but is also
possible in unstructured environments provided the terrain is sufficiently
textured to allow for repeatable keypoint identification and matching.}

\subsection{Observability Analysis and Regularization}
\label{sec:observability}

Underwater inspection vehicles are often rotationally stable by design, with
transitory roll and pitch excitation falling within $\pm 3 \deg$.  As a result,
patch test trajectories are largely planar, with some variation in vehicle
depth.  Unfortunately, the optimization problem defined by
\Cref{sec:reprojectionerrors} suffers a lack of observability under planar
vehicle motion.  This is revealed by expanding the Jacobian $\mbf{F}_j$ from
\eqref{eqn:extrinsics_perturb_epsilon},
\begin{equation}
    \mbf{F}_j = \begin{bmatrix}
        \big( \mbf{C}_1 \mbfbar{C} \, \mbf{r}_1^\times - \mbf{C}_2 \mbfbar{C} \, \mbf{r}_2^\times \big) & \colour{red}{\big( \mbf{C}_2 - \mbf{C}_1  \big)} \, \mbfbar{C}
    \end{bmatrix}.
    \label{eqn:extrinsics_F_expanded}
\end{equation}
Owing to the error definition and the peculiarity of the $(\cdot)^\odot$
operator \eqref{eqn:odot}, $\mbf{F}_j$ contains a \textit{difference in DCMs},
highlighted in red in \eqref{eqn:extrinsics_F_expanded}.  Under approximately
planar vehicle motion, the red terms in \eqref{eqn:extrinsics_F_expanded} become
\begin{equation}
    \colour{red}{\mbf{C}_2 - \mbf{C}_1} \approx \begin{bmatrix} \mbs{C}_2 & \mbs{\delta}_2 \\ \mbs{\delta}_2^\trans & 1 - \delta_2 \end{bmatrix} - \begin{bmatrix} \mbs{C}_1 & \mbs{\delta}_1 \\ \mbs{\delta}_1^\trans & 1 - \delta_1 \end{bmatrix} \approx \begin{bmatrix} \mbf{D} & \mbs{\delta} \\ \mbs{\delta}^\trans & \delta \end{bmatrix},
\end{equation}
where ${\mbs{C}_i \in SO(2), i=1,2}$ and with ${\mbs{\delta} \in \rnums^2}$,
$\delta$ some small values.  In a typical vehicle integration the Voyis Insight
Pro scanner is oriented directly downward, and the initial mean attitude
estimate $\mbftilde{C}_{b\ell}$ is simply an ENU-to-NED principal rotation.
This means the last three columns of $\mbf{F}_j$ are initially 
\begin{equation}
    (\mbf{C}_2 - \mbf{C}_1) \, \mbftilde{C} = \begin{bmatrix} \mbf{D} & \mbs{\delta} \\ \mbs{\delta}^\trans & \delta \end{bmatrix}
        \begin{bmatrix}
            0 & 1 & 0 \\ 1 & 0 & 0 \\ 0 & 0 & -1
        \end{bmatrix} = \begin{bmatrix}
            \star & -\mbs{\delta} \\ \star & -\delta
        \end{bmatrix},
\end{equation}
where the $\star$ entries are assumed to be full column rank.  Under
approximately planar vehicle motion $\mbs{\delta}$ and $\delta$ are small, the
full Jacobian matrix ${\mbf{F} = [ \, \begin{matrix} \mbf{F}_1^\trans & \cdots &
\mbf{F}_M^\trans \end{matrix} \, ]^\trans }$ no longer has full numerical column
rank, and component $\delta \rho_3$ of the state update will no longer be
observable.  This same observability issue is noted in
\cite{Mirzaei2012,Taylor2016,Zuo2020} for extrinsic calibration on ground
vehicles.

In contrast to the observability-aware update approach used in \cite{Lv2022a},
this work uses Tikhonov regularization \cite[\textsection6.3.2]{Boyd2004} to
ensure the state update $\delta \mbs{\xi}_\star$ remains a reasonable size when
$\mbf{F}$ is poorly conditioned.  Tikhonov regularization is analogous to
including a prior measurement ${\mbfcheck{T}_0 = \mbftilde{T}_0 \exp(-\delta
\mbs{\eta}_0^\wedge)}$ on the laser-to-INS extrinsics $\mbf{T}$, leading to the
objective function
\begin{equation}
    J_1(\mbf{T}) = \frac{1}{2} \Big( \| \mbf{e}_0(\mbf{T}) \|^2_{\mbf{P}_0\inv} + \sum^M_{j=1} \| \mbf{e}_j(\mbf{T}) \|^2_{\mbf{M}_j\inv} \Big).
    \label{eqn:extrinsics_obj_func_tik}
\end{equation}
The (left-invariant) prior error $\mbf{e}_0(\mbf{T})$ takes the form 
\begin{subequations}
    \begin{align}
        \mbfbar{e}_0 =& \ \log(\mbfbar{T}\inv \mbftilde{T}_0)^\vee, \\
        \mbf{e}_0 =& \ \mbfbar{e}_0 + \underbrace{\mbf{J}^\ell(\mbfbar{e}_0)\inv}_{\mbf{F}_0} \delta \mbs{\xi} \underbrace{- \mbf{J}^\textrm{r}(\mbfbar{e}_0)\inv}_{\mbf{G}_0} \delta \mbs{\eta}_0,
    \end{align}
    \label{eqn:extrinsics_batch_prior}%
\end{subequations}
where $\mbf{J}^\ell$ and $\mbf{J}^\textrm{r}$ are, respectively, the left and
right Jacobians of $SE(3)$ \cite[\textsection7.1.5]{Barfoot2017}, and where
parameter ${ \mbs{\Sigma}_0 = \expect{ \, \delta \mbs{\eta}_0 \, \delta
\mbs{\eta}_0^\trans \, } }$ is the covariance on the prior measurement, with
${\mbf{P}_0 = \mbf{G}_0 \mbs{\Sigma}_0 \mbf{G}_0^\trans}$.  This parameter is
tuned according to the certainty of the prior measurement, for example whether
the measurement is from a customer metrology report or is taken by hand in the
field.

Tikhonov regularization allows for reasonably-sized updates along poorly
observable dimensions.  This keeps the problem well-conditioned while still
allowing it to incorporate information from any residual pitch or roll
excitation that may be present in the vehicle trajectory estimate.  This
approach is different from the observability-aware update used in
\cite{Lv2022a}, which performs a truncated SVD to simply reject updates in
poorly observable dimensions.  In the current application, such an approach may
entail that the third component of $\mbf{r}^{sz}_b$ is never updated from its
prior value, despite weakly observable evidence in the calibration data that an
update is merited.

\subsection{Three Algorithms for Laser-to-INS Extrinsic Calibration}
\label{sec:threealgorithms}

Three algorithms for laser-to-INS extrinsic calibration are presented in this
section, subject to increasingly weaker assumptions on the quality of the
vehicle trajectory estimate (\Cref{tab:algs_overview}).  This allows for good
extrinsic calibration in a wide range of practical scenarios.

\vspace{3pt}
\subsubsection{Algorithm 1: Perfect Navigation}
\label{sec:alg1}

Algorithm 1 assumes the vehicle navigation is perfect, and only optimizes over
the laser-to-INS extrinsics $\mbf{T}^{sz}_{b\ell}$.  This assumption could be
valid in scenarios where high-precision navigational aiding is available, for
example a long-baseline (LBL) acoustic array, or GPS in surface applications.
The result is a nonlinear least-squares optimization problem of the form
\eqref{eqn:optprob}, with $J(\mbc{X})$ simply replaced by the objective function
$J_1(\mbf{T})$ from \eqref{eqn:extrinsics_obj_func_tik}.

\vspace{3pt}
\subsubsection{Algorithm 2: Good Local Navigation with Global Drift}
\label{sec:alg2}

\begin{figure}[t]
	\centering
	\includegraphics[width=\columnwidth]{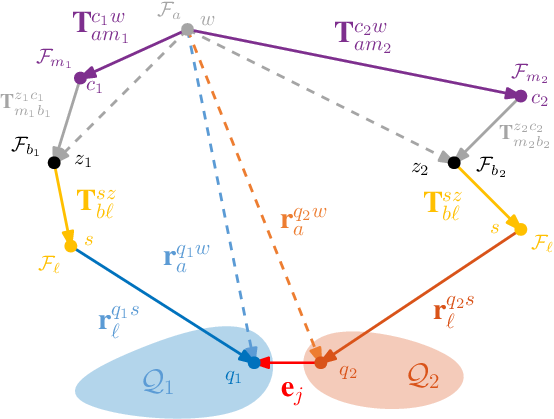}
	\caption{Defining a reprojection error \colour{red}{$\mbf{e}_j$} for
	Algorithm 2, which allows for global submap drift.  Comparison to
	\Cref{fig:reprojection_errors_1} shows the individual vehicle poses (black)
	have been replaced with the \colour{matlab_purple}{central submap poses} and
	\colour{voyis_grey}{rigid offsets}.  The design variables are the
	\colour{voyis_yellow}{laser-to-INS extrinsics} and the
	\colour{matlab_purple}{central submap poses}.}
    \label{fig:reprojection_errors_2}
\end{figure}

A more realistic assumption for underwater navigation is good local navigation
with global drift.  In this scenario an underwater vehicle is equipped with a
high-quality DVL-INS, and either dead-reckons or receives intermittent
correction from a lower-precision acoustic sensor such as a surface-mounted
ultrashort baseline (USBL) array.  The resulting point-cloud submaps are assumed
to be locally rigid, but are allowed some degree of global pose refinement to
correct for gross navigation errors.  Algorithm 2 accounts for this by including
the central poses of the $N$ submaps as design variables, with a prior term to
control the degree of global pose correction.

Consider \Cref{fig:reprojection_errors_2}, where the reprojection error
$\mbf{e}_j$ is redefined in terms of both the laser-to-INS extrinsics
$\mbf{T}^{sz}_{b\ell}$, shown in yellow, and the central submap poses
$\mbf{T}^{cw}_{am}$, shown in purple.  \update{The central poses are chosen to
be the poses in each submap that lie nearest, in a Euclidean sense, to the
identified trajectory crossing.  For readability, the central poses are
identified by the vehicle datum $c$ and the vehicle reference frame
$\rframe{m}$.}  The relative poses $\mbf{T}^{zc}_{mb}$, shown in grey in
\Cref{fig:reprojection_errors_2}, parameterize each vehicle pose within a submap
relative to the central submap pose at the time each keypoint was measured by
the laser.  To ensure the submap remains rigid, these relative poses are
precomputed and are treated as parameters in the optimization.  

To derive the the ingredients needed for batch optimization
\eqref{eqn:gauss_newton}, first use \Cref{fig:reprojection_errors_2} to redefine
the reprojection error as 
\begin{subequations}
    \begin{align*}
        \mbf{e}_j(\mbf{T}^{sz}_{b\ell}, \mbf{T}^{c_1w}_{am_1}, \mbf{T}^{c_2w}_{am_2})        =& \ \mbf{r}^{q_1w}_{a} - \mbf{r}^{q_2w}_{a} \\
        \begin{split}
            =& \ \mbf{H} \Big( \mbf{T}^{c_1w}_{am_1} \, \mbf{T}^{z_1c_1}_{m_1b_1} \, \mbf{T}^{sz}_{b\ell} \, \mbf{u}^{q_1s}_\ell \\
            & \ - \mbf{T}^{c_2w}_{am_2} \, \mbf{T}^{z_2c_2}_{m_2b_2} \, \mbf{T}^{sz}_{b\ell} \, \mbf{u}^{q_2s}_\ell \Big),
        \end{split}
        \numberthis \label{eqn:rep_error_joint_1} \\
        \mbf{e}_j(\mbf{T}, \mbf{T}_1, \mbf{T}_2) =& \ \mbf{H} \big( \mbf{T}_1 \, \mbftilde{T}_1 \mbf{T} \, \mbf{u}_1 - \mbf{T}_2 \, \mbftilde{T}_2 \mbf{T} \, \mbf{u}_2 \big),
        \numberthis \label{eqn:rep_error_joint_2}
    \end{align*}
\end{subequations}
in which the homogenous form $\mbf{u}$ is used for the point measurements and
where the notation is simplified from \eqref{eqn:rep_error_joint_1} to
\eqref{eqn:rep_error_joint_2} such that, for example, ${\mbf{T}_1 \leftarrow
\mbf{T}^{c_1w}_{am_1}}$, ${\mbftilde{T}_1 \leftarrow
\mbf{T}^{z_1c_1}_{m_1b_1}}$, ${\mbf{T} \leftarrow \mbf{T}^{sz}_{b\ell}}$, and
${\mbf{u}_1 \leftarrow \mbf{u}^{q_1s}_\ell}$.  Perturbing design variables
$\mbf{T}$, $\mbf{T}_1$, and $\mbf{T}_2$ in a left-invariant sense, with
${\exp(-\delta \mbs{\xi}^\wedge) \approx (\eye - \delta \mbs{\xi}^\wedge)}$,
\eqref{eqn:rep_error_joint_2} is linearized as 
\begin{align*}
    \mbf{e}_j =& \ \mbfbar{e}_j + \underbrace{ \mbf{H} \big( \mbfbar{T}_2 \, \mbftilde{T}_2 \mbfbar{T} \, \mbfbar{u}_2^\odot - \mbfbar{T}_1 \, \mbftilde{T}_1 \mbfbar{T} \, \mbfbar{u}_1^\odot \big)}_{\mbf{F}_j} \delta \mbs{\xi} \\
    & \   \underbrace{-\mbf{H} \big( \mbfbar{T}_1 \big( \mbftilde{T}_1 \mbfbar{T} \, \mbfbar{u}_1 \big)^\odot \big)}_{\mbf{F}^1_j} \delta \mbs{\xi}_1 + \underbrace{ \mbf{H} \big( \mbfbar{T}_2 \big( \mbftilde{T}_2 \mbfbar{T} \, \mbfbar{u}_2 \big)^\odot \big)}_{\mbf{F}^2_j} \delta \mbs{\xi}_2 \\
    & \ + \underbrace{\mbfbar{C}_1 \mbftilde{C}_1 \mbfbar{C}}_{\mbf{G}^1_j} \delta \mbf{r}_1 \underbrace{-\mbfbar{C}_2 \mbftilde{C}_2 \mbfbar{C}}_{\mbf{G}^2_j} \delta \mbf{r}_2,
    \numberthis \label{eqn:jac_rep_2}
\end{align*}
where second-order terms are again ignored.  Following the same substitutions
and assumptions used in \Cref{sec:reprojectionerrors}, the covariance on the
reprojection error is given by \eqref{eqn:extrinsics_sigma}.  Incorporating the
Tikhonov regularization term from \eqref{eqn:extrinsics_batch_prior} and placing
prior measurements on each of the $N$ submaps, the objective function for
Algorithm 2 becomes
\begin{equation}
    J_2(\mbc{X}) \! = \! \frac{1}{2} \Big(  \! \| \mbf{e}_0(\mbf{T}) \|^2_{\mbf{P}_0\inv} \hspace{-0.5pt} + \hspace{-0.5pt} \sum^N_{i=1} \hspace{-0.5pt} \| \mbf{e}_i(\mbf{T}_i) \|^2_{\mbf{P}_i\inv} \hspace{-0.5pt} + \hspace{-0.5pt} \sum^M_{j=1} \hspace{-0.5pt} \| \mbf{e}_j(\mbc{X}_j) \|^2_{\mbf{M}_j\inv} \! \hspace{-1pt} \Big),
    \label{eqn:extrinsics_obj_func_2}
\end{equation}
where the design variables are ${\mbc{X} = \{ \mbf{T}^{sz}_{b\ell},
\mbf{T}^{c_1w}_{am_1}, \ldots, \mbf{T}^{c_Nw}_{am_N} \} }$, and where
user-defined parameter $\mbs{\Sigma}_i$ controls the strength of the global
submap priors, with ${\mbf{P}_i = \mbf{G}_i \mbs{\Sigma}_i \mbf{G}_i^\trans}$
(see \eqref{eqn:extrinsics_batch_prior}).

\vspace{3pt}
\subsubsection{Algorithm 3: Poor Navigation}
\label{sec:alg3}

Algorithm 3 assumes the local vehicle trajectory estimate is of poor quality,
for example when the vehicle is equipped with a MEMS-based IMU.  In this
scenario the rigid submap assumption used in Algorithm~2 may no longer be valid,
leading to the general case of laser-to-INS extrinsic calibration using flexible
submaps.  

\begin{figure}[t]
	\centering
	\includegraphics[width=\columnwidth]{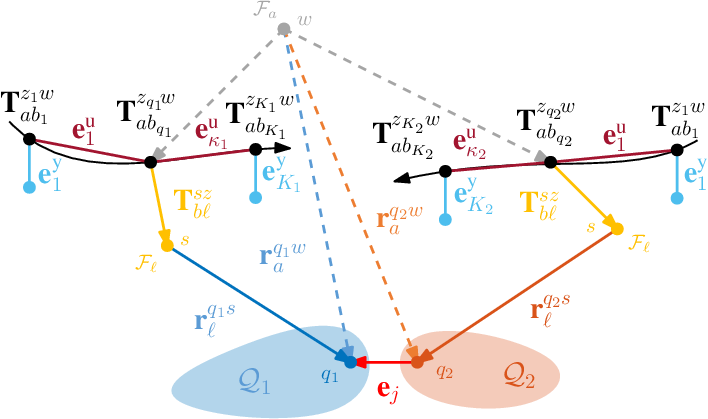}
	\caption{A depiction of Algorithm 3, with \colour{matlab_cyan}{exteroceptive errors}
	placed on the DVL-INS poses and \colour{matlab_maroon}{interoceptive errors}
	linking all DVL-INS and measurement poses, both shown in black.  The design
	variables include the \colour{voyis_yellow}{laser-to-INS extrinsics} as well
	as the \colour{matlab_maroon}{intrinsic} and \colour{matlab_cyan}{extrinsic}
	variables.}
    \label{fig:reprojection_errors_3}
    \vspace{-9pt}
\end{figure}

Consider \Cref{fig:reprojection_errors_3}, where the $i^\textrm{th}$ flexible
submap is constructed using $K_i$ vehicle poses, ${\mbf{T}_k =
\mbf{T}^{z_kw}_{ab_k}, k = 1, \ldots, K_i}$.  The submap is globally constrained
by the exteroceptive error terms $\mbf{e}^{\textrm{y}}$, but is allowed to
adjust its shape through the interoceptive error terms $\mbf{e}^{\textrm{u}}$
linking adjacent poses.  This adaptability will be needed in situations where
the short-term navigation drift within each submap is no longer negligible, for
example when using low-cost underwater vehicles.

The experimental results in this work use data from a DVL-INS, which does not
provide access to raw interoceptive measurements but instead provides pose
measurements ${\mbfcheck{T}_k \in SE(3)}$.  As a result, this work assumes a
white-noise-on-acceleration (WNOA) vehicle motion prior \cite{Anderson2015},
leading to WNOA interoceptive errors of the form 
\vspace{-3pt}
\begin{equation}
    \mbf{e}_k^{\textrm{w}} = \log \Big( \mbf{T}_k\inv \big( \mbf{T}_{k-1} \exp \big( \delta t_{k-1} \mbs{\varpi}^\wedge_{k-1} \big) \big) \Big)^\vee,
    \vspace{-3pt}
\end{equation}
with $\delta t$ a time increment and ${\mbs{\varpi} \in \rnums^6}$ the
generalized velocity.  See \cite{Hitchcox2023a} for a comprehensive derivation.
Additionally, relative pose errors of the form 
\vspace{-3pt}
\begin{equation}
    \mbf{e}_k^{\textrm{r}} = \log \left( \mbf{T}_k\inv \left( \mbf{T}_{k-1} \mbfcheck{T}\inv_{k-1} \mbfcheck{T}_k \right) \right)^\vee
    \vspace{-3pt}
\end{equation}
were found in \cite{Hitchcox2023a} to retain some rigidity throughout the
submaps, allowing global corrections to propagate more easily.  Finally, prior
pose errors of the form 
\vspace{-3pt}
\begin{equation}
    \mbf{e}^{\textrm{p}}_k = \log \big( \mbf{T}_k\inv \mbfcheck{T}_k \big)^\vee
    \vspace{-3pt}
\end{equation}
are constructed for each vehicle pose using the DVL-INS measurements.
Incorporating reprojection errors $\mbf{e}_j$, the prior pose errors
$\mbf{e}^{\textrm{p}}_k$ and interoceptive errors $\mbf{e}_k^{\textrm{w}}$ and $\mbf{e}_k^{\textrm{r}}$ from
above, and the Tikhonov regularization term from
\eqref{eqn:extrinsics_batch_prior}, the objective function for Algorithms 3 is 
\vspace{-3pt}
\begin{equation}
    \begin{split}
        J_3(\mbc{X}) =& \ \frac{1}{2} \Big( \sum^M_{j=1} \| \mbf{e}_j(\mbc{X}_j) \|_{\mbf{M}_j\inv}^2 + \sum^N_{i=1} \Big( \sum^{K_i}_{k=1}  \| \mbf{e}^{\textrm{p}}_k(\mbf{T}_k) \|^2_{\mbf{P}_k\inv} \\ 
                      & \hspace{-14pt} + \sum^{\kappa_i}_{k=1} \Big( \| \mbf{e}^{\textrm{w}}_k(\mbc{X}^{\textrm{w}}_k) \|^2_{\mbf{Q}_k\inv} + \| \mbf{e}^{\textrm{r}}_k(\mbc{X}^{\textrm{r}}_k) \|^2_{\mbf{R}_k\inv} \Big) \Big) + \| \mbf{e}_0 \|^2_{\mbf{P}_0\inv} \Big).
    \end{split}
    \label{eqn:j3}%
\end{equation}
Unpacking \eqref{eqn:j3}, the prior pose error covariance $\mbf{P}_k$ is
generated from the DVL-INS measurement covariance $\mbs{\Sigma}_k$ (see
\eqref{eqn:extrinsics_batch_prior}), while covariances $\mbf{Q}_k$ and
$\mbf{R}_k$ incorporate user-defined parameters (see \cite{Hitchcox2023a}).  The
limit ${\kappa_i = K_i+Q_i-1}$ of the last sum reflects the fact that the
interoceptive errors must incorporate both the DVL-INS vehicle poses $\mbf{T}_k$
as well as the keypoint measurement poses $\mbf{T}_q$, with $Q_i$ the total
number of keypoints detected in the $i^\textrm{th}$ submap.  The set of design
variables $\mbc{X}$ in Algorithm 3 is quite large, incorporating the
laser-to-INS extrinsics, the vehicle poses at the DVL-INS timestamps $t_k$ as
well as the keypoint timestamps $t_q$, and all intervening generalized
velocities.

\section{Results}
\label{sec:results}

The developed algorithms are used for laser-to-INS extrinsic calibration on two
experimental datasets: a small shipwreck dataset collected by a surface vessel,
and a laser model of the \textit{Endurance} shipwreck collected by a SAAB
Sabertooth AUV during the Endurance22 expedition \cite{Rabenstein2022}.

\subsection{Experimental Results: Wiarton Shipwreck}
\label{sec:expresults}

\begin{figure}[b]
	\sbox\subfigbox{%
	  \resizebox{\dimexpr\columnwidth-1em}{!}{%
		\includegraphics[height=3cm]{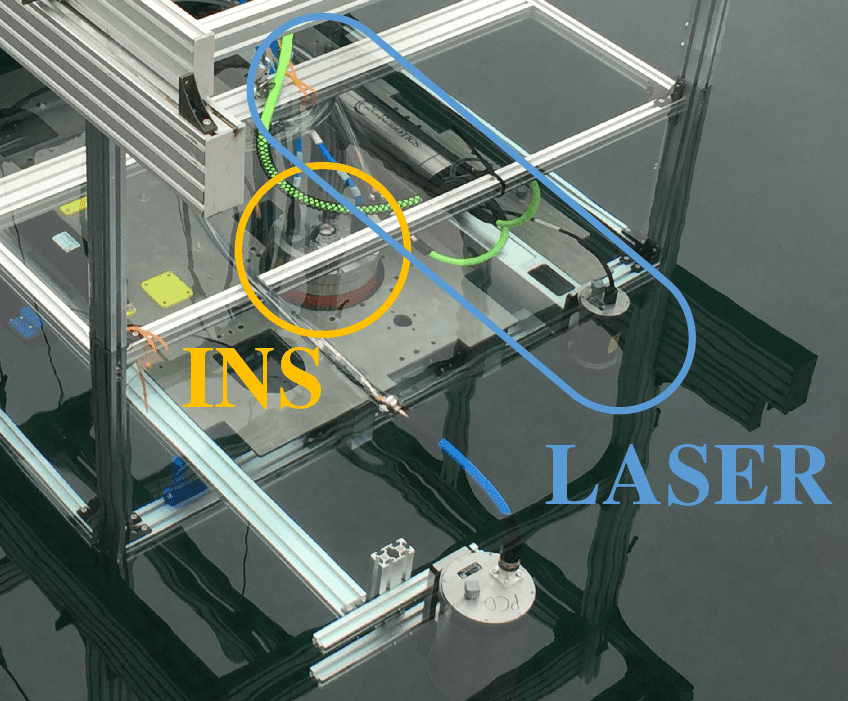}%
		\includegraphics[height=3cm]{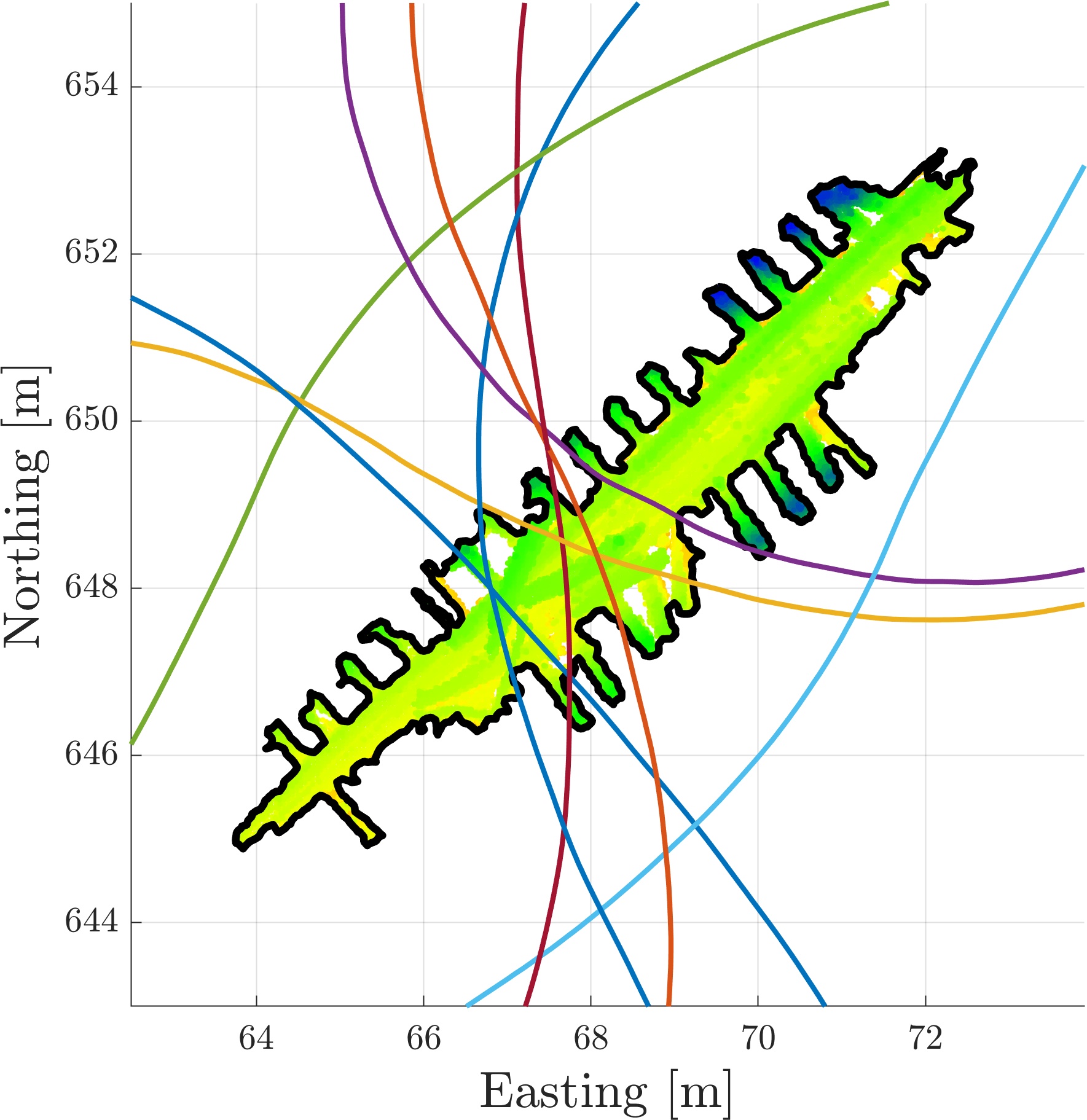}%
	  }%
	}
	\setlength{\subfigheight}{\ht\subfigbox}
	\centering
	\subcaptionbox{Sensor payload \label{fig:payload}}{%
	  \includegraphics[height=\subfigheight]{figs/voyis_payload_labels_crop.png}
	}%
    \hspace{3pt}%
	\subcaptionbox{Wreck and passes \label{fig:wiarton_wreck}}{%
	  \includegraphics[height=\subfigheight]{figs/trajectories_zoom.jpg}
	} 
    \caption{Field testing with Voyis Imaging Inc.  The surface vessel payload
	is shown in \Cref{fig:payload}, while the shipwreck structure is shown in
	\Cref{fig:wiarton_wreck}.  Individual vehicle passes over the wreck area are
	highlighted in different colours. }
  \label{fig:exp_1_deployment}
\end{figure}

In this section, all three algorithms are used for laser-to-INS extrinsic
calibration on a surface vessel.  \Cref{fig:payload} shows the sensor payload on
the bow of the vessel, with a Sonardyne SPRINT-Nav DVL-INS and Voyis Insight Pro
laser scanner highlighted.  The vessel scanned a small shipwreck in shallow
water during a field deployment to Colpoy's Bay, located in Wiarton, Ontario,
Canada.  \Cref{fig:wiarton_wreck} shows the main shipwreck structure and eight
sections of the vehicle trajectory.  The full details of this field deployment
may be found in \cite{Hitchcox2023a}.  

A total of ${N=8}$ point-cloud submaps are first constructed from a GPS-aided
DVL-INS trajectory estimate.  Reprojection errors $\mbf{e}_j$ are formed between
the submaps from inlier keypoint matches identified by TEASER++, and extrinsic
optimization is performed using Algorithms~1-3.  The initial mean laser-to-INS
extrinsic estimate $\mbftilde{T}^{sz}_{b\ell}$ is obtained from a CAD model of
the sensor rig, and the prior measurement covariance is set to 
\begin{equation}
    \mbs{\Sigma}_0 = \blkdiag( \sigma^\phi_0 \, \eye, \, \sigma^\rho_0 \, \eye)^2,
    \label{eqn:sigma0_structure}
\end{equation}
with ${\sigma^\phi_0 = \SI{1}{\deg}}$ and ${\sigma^\rho_0 =
\SI{5}{\centi\meter}}$.  The global submap pose covariance $\mbs{\Sigma}_i$ used
in Algorithm 2 follows the same parameter structure, with ${\sigma^\phi_i =
\SI{1}{\deg}}$ ${\sigma^\rho_i = \SI{25}{\centi\meter}}$.

Updates to the laser-to-INS extrinsics are reported as
\begin{subequations}
    \begin{align}
        \delta \mbs{\phi}_{\ell b} =& \ \log( \mbftilde{C}^\trans \mbf{C}_\star )^\vee, \\
        \delta \mbf{r}^{sz}_b =& \ \mbf{r}_\star - \mbftilde{r}.
        \label{eqn:extrinsics_compare_to_cad_displacement}
    \end{align}
    \label{eqn:extrinsics_compare_to_cad}%
\end{subequations}
Reporting the updates on ${SO(3) \times \rnums^3}$ means the position updates
are resolved in the body frame of the vehicle, which is easier to interpret.
Updates produced by the three algorithms are reported in
\Cref{tab:extrinsics_wiarton}.  

\begin{figure}[t]
	\centering
	\includegraphics[width=\columnwidth]{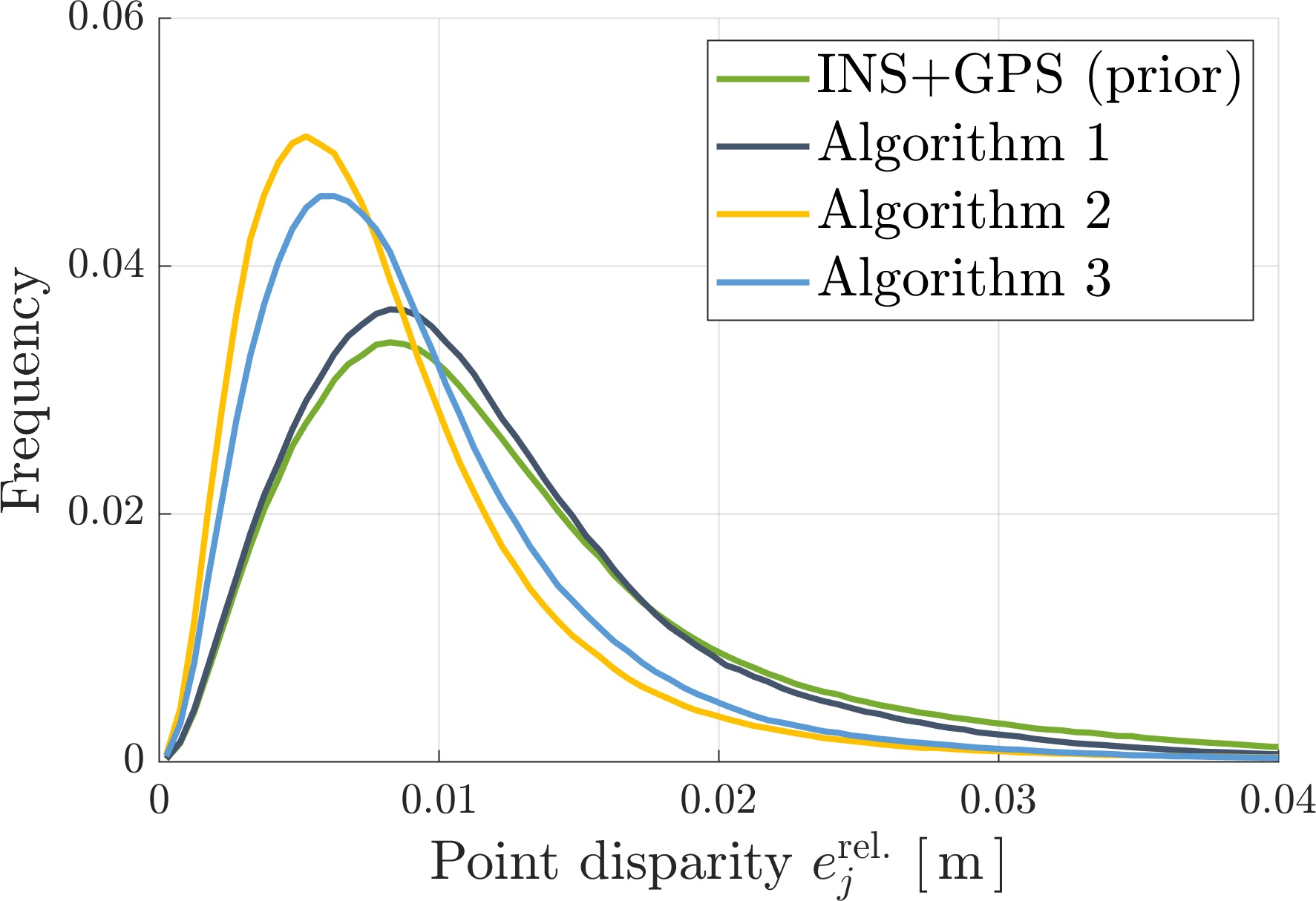}
	\caption{Empirical PDFs on the point disparity error for the Wiarton
	shipwreck dataset.  The ``INS+GPS'' result is from \cite{Hitchcox2023a},
	while Algorithms 1-3 are the methods from \Cref{sec:threealgorithms}.}
    \label{fig:wiarton_pdf}
\end{figure}

\begin{table}[t]
    \centering
    \caption{Change from CAD values for Wiarton extrinsics}
    \renewcommand{\arraystretch}{1.5}
    \begin{tabularx}{\columnwidth}{C{0.75cm}|Y|YYY}
    \toprule
    Alg. & $\| \delta \mbs{\phi} \| \, \left[ \deg \right] $ & $\delta r^1_b \,
    \left[ \SI{}{\centi \meter} \right]$ & $\delta r^2_b \, \left[ \SI{}{\centi
    \meter} \right]$ & $\delta r^3_b \, \left[ \SI{}{\centi \meter} \right]$ \\
    \hline
        1 & \SI{0.17}{} & \SI{0.77}{} & -3.44 & -3.80 \\
        2 & \SI{0.18}{} & -\SI{3.20e-3}{} & -\SI{0.32}{} & \SI{4.10e-3}{} \\
        3 & \SI{0.18}{} & -\SI{6.89e-2}{} & -1.92 & \SI{7.27e-2}{} \\
	\bottomrule
    \end{tabularx}
	\label{tab:extrinsics_wiarton}
\end{table}

Results are also assessed by computing the \textit{point disparity errors}
\cite{Roman2006} in the posterior point-cloud submaps.  For each point, the
point disparity error is simply the Euclidean distance to the closest point in
any of the other ${N-1}$ submaps, with a low average disparity error indicating
a well-aligned, ``crisp'' point-cloud map.  \Cref{fig:wiarton_pdf} shows
empirical probability density functions (PDFs) for the three calibration
algorithms as well as for the prior GPS-aided ``INS+GPS'' trajectory from
\cite{Hitchcox2023a}.

Together, the results make a strong case for the applicability of Algorithm~2 in
this particular context.  Algorithm~2 is the most successful in reducing the
point disparity error, as seen by comparing the yellow and green curves in
\Cref{fig:wiarton_pdf}.  Additionally, the small updates suggested by
Algorithm~2 in \Cref{tab:extrinsics_wiarton} are consistent with mounting
tolerances for a CAD-designed, machined plate (\Cref{fig:payload}).  The
improved performance of Algorithm~2 over Algorithm~1, which assumes a perfect
navigation estimate, suggests a possible bias in the post-processed GPS data,
perhaps from an error in the GPS-to-INS extrinsic estimate.  \update{The
improved performance of Algorithm~2 over Algorithm~3 suggests the minimization
of the cost $J_3(\mbc{X})$ from \eqref{eqn:j3} favoured the reduction of the
squared prior errors $\mbf{e}^\textrm{p}_k$ and WNOA errors
$\mbf{e}^\textrm{w}_k$ over the reduction of the squared point disparity errors
$\mbf{e}_j$.  Further tuning of the $\mbf{P}_k$ and $\mbf{Q}_k$ covariance
matrices may lead to improved performance for Algorithm~3 in scenarios involving
low-cost navigation systems.}


\subsection{Experimental Results: Wreck of the \textit{Endurance}}
\label{sec:endurance}

The Endurance, captained by Sir Ernest Shackleton, sank in 1915 during an
ill-fated attempt to cross Antarctica. The story that followed is a well-known
case study in stoicism and leadership, whereby Shackleton and all 27 members of
his crew survived following an epic trial of ocean navigation
\cite{Lansing2014}. The ship was not as fortunate, and sank in the Weddell Sea
on Nov. 21, 1915 after being crushed by ice.  Given the harsh and extremely
remote environment, the exact location of the wreck remained a mystery for over
one hundred years.

The Endurance22 expedition, organized by the Falklands Maritime Heritage Trust,
was launched in Feb. 2022 with the primary goal of locating the wreck of the
\textit{Endurance}.  Once located, a SAAB Sabertooth AUV was used to inspect the
wreck site.  The vehicle was equipped with a Sonardyne SPRINT-Nav DVL-INS, a
USBL beacon, a depth sensor, and an Insight Pro laser line scanner from Voyis
Imaging Inc.

The wreck site was located at a depth of $\SI{3}{\kilo\meter}$.  At this range,
USBL signals from the surface vessel were not precise enough to aid navigation,
and the thick layer of sea ice made the deployment of an LBL array impossible.
The AUV trajectory was therefore dead-reckoned for the duration of the
inspection.  

As part of the efforts to construct a 3D point-cloud model of the wreck site,
Algorithm 2 is used to calibrate the laser-to-INS extrinsics using a dedicated
patch test dataset with ${N=7}$ submaps.  The assumptions behind Algorithm 2
make sense in this context, where high-quality dead-reckoning is available and
localizing measurements are unavailable.  

\begin{figure}[t]
	\centering
	\includegraphics[width=\columnwidth]{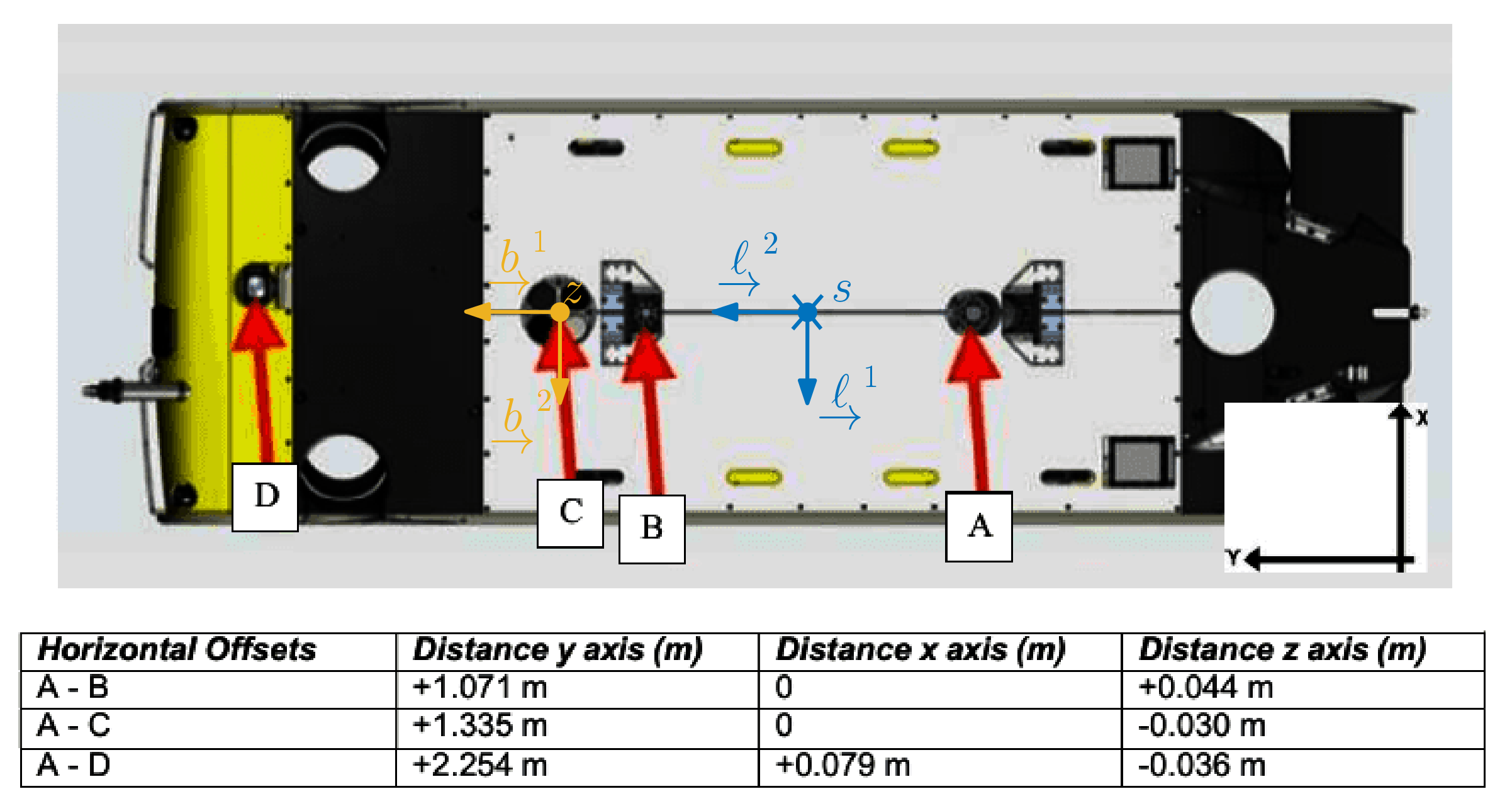}
	\caption[The metrology report for the SAAB Sabertooth AUV used on the
	Endurance22 mission, modified to show the pose of the laser relative to the
	INS]{A metrology report for the SAAB Sabertooth AUV used on the Endurance22
	expedition, modified to show the \colour{matlab_blue}{pose of the laser}
	relative to the \colour{matlab_yellow}{DVL-INS}.  Modified with permission
	from the Falklands Maritime Heritage Trust.}
	\label{fig:extrinsics_metrology_report}
\end{figure}

A good initial guess is provided by the AUV metrology report, shown in
\Cref{fig:extrinsics_metrology_report}.  Though this report was generated using
high-precision photogrammetry, the laser datum is inaccessible behind the AUV
housing and its exact location is somewhat unclear.  The prior displacement
measurement $\mbftilde{r}^{sz}_{b}$ is taken to lie midway between points \texttt{A}
and \texttt{B} in \Cref{fig:extrinsics_metrology_report}, resulting in 
\begin{equation}
    \mbftilde{r}^{sz}_b = \begin{bmatrix}
        -0.80 & 0 & 0
    \end{bmatrix}^\trans,
\end{equation}
while an ENU-to-NED principal rotation is taken as the prior attitude
measurement $\mbftilde{C}_{b\ell}$.  The parameter values from
\eqref{eqn:sigma0_structure} for the prior extrinsic and global submap pose
measurements are ${\sigma_0^\phi = \sigma_i^\phi = \SI{1}{\deg}}$,
${\sigma^\rho_0 = \SI{0.1}{\meter}}$, and ${\mbs{\Sigma}^\rho_i = \diag(1^2,
1^2, (0.1)^2) \, \SI{}{\meter}}$.  The non-isotropic structure of
$\mbs{\Sigma}^\rho_i$ reflects greater precision in depth owing to the
availability of depth sensor measurements.  

After running Algorithm 2, the update to the laser-to-INS extrinsics is computed
using \eqref{eqn:extrinsics_compare_to_cad}.  A reasonable value of ${\| \delta
\mbs{\phi}_{\ell b} \| = \SI{0.36}{\deg} }$ is obtained for the attitude update,
while the position update in centimeters is found to be 
\begin{equation}
    \delta \mbf{r}^{sz}_b = \begin{bmatrix}
        4.68 & \SI{-2.52e-3}{} & \SI{0.14}{}
    \end{bmatrix}^\trans.
    \label{eqn:extrinsics_saab_update}
\end{equation}
Considering \Cref{fig:extrinsics_metrology_report}, the displacement update lies
almost perfectly along $\ura{b}^1$, the main axis of the AUV.  The relative
position measurement in this dimension was initially unclear, as the laser datum
is inaccessible behind the AUV housing.  The displacement update makes sense,
indicating the scanner is \SI{4.68}{\centi \meter} closer to the DVL-INS than
initially assumed.

Finally, the quality of the posterior point-cloud map is assessed by computing
the point disparity errors for the patch test submaps.  The results are
visualized as a colourmap in \Cref{fig:extrinsics_endurance}.  Algorithm 2,
which jointly optimizes over both the laser-to-INS extrinsics and the global
submap poses, has reduced the median point disparity error from
$\SI{7.7}{\centi\meter}$ to $\SI{0.6}{\centi\meter}$.  The posterior point-cloud
map provides a strikingly crisp, high-resolution reconstruction of this historic
wreck.

\section{Conclusion}
\label{sec:conclusion}

Underwater laser scanners such as the Voyis Insight Pro are increasingly used
for high-resolution infrastructure inspection and environmental monitoring.
However, reliable laser-to-INS extrinsic calibration remains a challenge on
commercial surveys.  This work developed three novel algorithms for laser-to-INS
calibration using naturally occurring features.  All algorithms employ Tikhonov
regularization to address low-observability scenarios frequently encountered in
practice.  Each algorithm makes a different assumption on the quality of the
vehicle trajectory estimate, however Algorithm 2, which assumes good local
navigation with global drift, proved a good choice for both field datasets.  All
three algorithms were successfully used on a small shipwreck dataset from
Wiarton, Ontario, while Algorithm 2 was used to refine the laser-to-INS
extrinsics for the SAAB Sabertooth AUV used on the Endurance22 expedition.
Future work will focus on extrinsic calibration for low-cost systems, and on an
iterative approach to address instances of poor initialization.

\section*{Acknowledgment}

The authors would like to sincerely thank the Falklands Maritime Heritage Trust
for organizing the Endurance22 expedition, and for allowing this initial glimpse
of field data to appear in publication.  The authors would also like to thank
Nico Vincent and Pierre Legall at Deep Ocean Search for their wonderful support
and advice on this project.  Thanks to Ocean Infinity for conducting the survey,
and to SAAB for providing the AUV and performing sensor integration.  Thanks as
well to Voyis Imaging Inc. for their financial and technical support, and to
Sonardyne International Limited for their navigation expertise and data
processing.  Finally, thanks to Lasse Rabenstein at Drift+Noise Polar Services
for his role in the Endurance22 expedition, and for preparing such a
comprehensive summary report.

\vfill

\printbibliography


\end{document}